\documentclass[final,3p]{elsarticle}
\usepackage{amssymb}
\usepackage{amsmath}
\usepackage{xcolor}
\usepackage{setspace}
\usepackage{pdfpages}
\usepackage{caption}
\usepackage{textcomp}
\usepackage{subfigure}

\let\today\relax
\makeatletter
\def\ps@pprintTitle{%
	\let\@oddhead\@empty
	\let\@evenhead\@empty
	\def\@oddfoot{\footnotesize\itshape
		{} \hfill\today}%
	\let\@evenfoot\@oddfoot
}
\makeatother

\begin{document}

\begin{frontmatter}



\title{
The applicability of transperceptual and deep learning approaches to the study and mimicry of complex cartilaginous tissues
}


\author[eng]{Waghorne J.}
\author[music]{Howard C.}
\author[lux]{Hu H.}
\author[lux]{Pang J.}
\author[chem]{Peveler W.J.}
\author[music]{Harris L.}
\author[eng,mat]{Barrera O.\corref{cor1}}
\ead{olga.barrera@eng.ox.ac.uk}

\cortext[cor1]{Corresponding author}
 
%\cortext[cor1]

%{obarrera@brookes.ac.uk, olga.barrera@eng.ox.ac.uk}

\address[eng]{School of Engineering Computing and Mathematics, Oxford Brookes University}
\address[music]{School of Culture and Creative Arts, University of Glasgow}
\address[lux]{Department of Computer Science, University of Luxembourg}

\address[chem]{School of Chemistry, University of Glasgow}
\address[mat]{Department of Engineering Science, University of Oxford}

\begin{abstract}
Complex soft tissues, for example the knee meniscus, play a crucial role in mobility and joint health, but when damaged are incredibly difficult to repair and replace. This is due to their highly hierarchical and porous nature which in turn leads to their unique mechanical properties providing in joint stability, load redistribution and friction reduction. 
In order to design tissue substitutes, the internal architecture of the native tissue needs to be understood and replicated.
Here we explore a combined audio-visual approach - so called transperceptual -  to generate artificial architectures mimicking the native ones. The proposed method uses both traditional imagery, and sound generated from each images as a method of rapidly comparing and contrasting the porosity and pore size within the samples. We have trained and tested a generative adversarial network (GAN) on the 2D image stacks. 
In order to understand how the resolution of the training set of images impacts on similarity of the artificial dataset to the original one, we have trained the GAN with two datasets. The first consisting of n=478 pairs of audio and image files for which the images were downsampled to 64 $\times$ 64 pixels, the second one consisting of n=7640 pairs of audio and image files for which the full resolution 256 $\times$ 256 pixels is retained but each images is divided into 16 squares to maintain the limit of 64 $\times$ 64 pixels required by the GAN.
We reconstruct the 2D stacks of artificially generated datasets into 3D objects and run image analysis algorithms to characterize statistically the architectural parameters - pore size, tortuosity and pore connectivity - and compare them with the original dataset. 
Results show that the artificially generated dataset that undergoes downsampling performs better in terms of parameter matching - values ranged within 4-8\% of that of the native tissues with the parameters: mean luminance of images, mean porosity and pore size. Our audiovisual approach has the potential to be extended to larger data sets in order to explore both how similarities and differences can be audibly recognised across multiple samples.

\end{abstract}

\begin{keyword}
hierarchical porous media \sep audio-visual \sep sonification \sep machine learning \sep meniscus
  


\end{keyword}

\end{frontmatter}

\section{Introduction}

Natural materials, such us soft tissues, present spatially inhomogeneous architectures often characterized by a hierarchical distribution of pores~\citep{agustoni2021high,VetrietAl,ELMUKASHFI2022425, maritz2020development, maritz2021functionally, bonomo2020procedure}. Due to the shape of pores observed at different scales and the resulting intricate network of pore connectivity, the characterization of architectural parameters - porosity, pore size, pore connectivity and tortuosity - of these objects is a non trivial task and it is object of an extensive research effort~\citep{Rabbani2021, COOPER20141033, AN2016156}.
It has been observed that the knee meniscus is one such example of a soft tissue with remarkable properties in terms of the ability of load transfer from the upper to lower part of the body. The structure is similar to a sandwich structure with a stiff outside layer and a softer internal layer so that it can accommodate deformation and dissipate energy. This tissue can be seen as an effective damping system designed and optmised by nature ~\citep{WaghornePerm}. The secret of the internal layer is the arrangement of the collagen in a non uniform network of channels oriented in a preferential direction guiding the fluid flow paths~\citep{VetrietAl,WaghornePerm, WaghorneTort}. Understanding how mechanical parameters such us permeability is linked to the morphology of these channels and their interconnection/tortuosity is essential to design suitable biomimetic systems that can be adopted for replacement or repair of meniscal injury, and to date there are no such suitable artificial material systems.

\begin{figure}[tbp]
\centering
\includegraphics[scale=1]{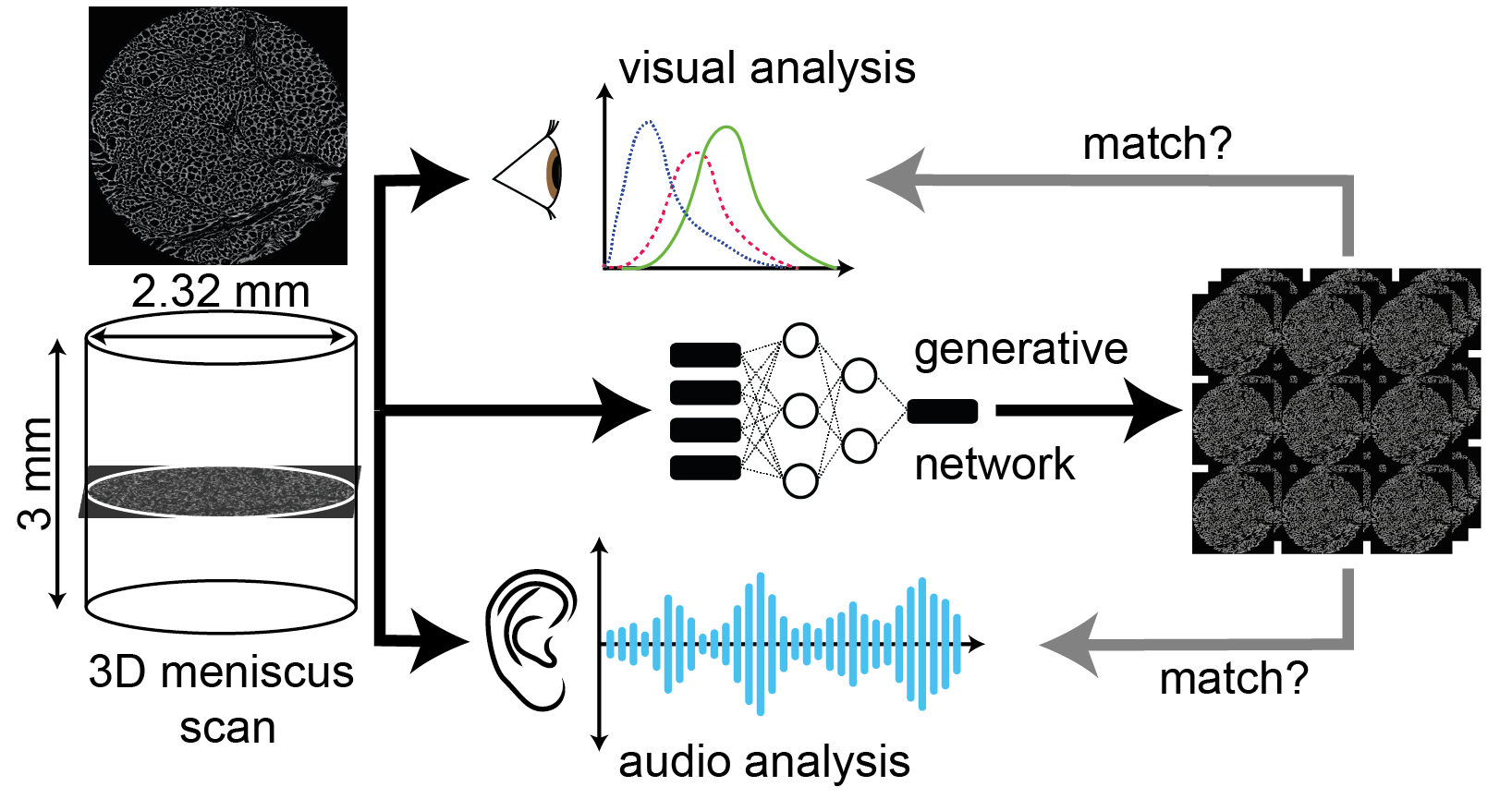}
\caption{Schematic outline of our transperceptual approach where 3D scans of meniscal tissue have been processed with both 'classical' visual approaches and sonification, and a generative network has been used to explore inter-conversion of the data types and whether generative architectures can mimic real measurements, as measured by the accuracy of image recovery from audio data.}
\label{fig:scheme}
\end{figure}

Designing artificial structures which can resemble both the internal architecture and the material properties of such natural objects is currently an area of ongoing study~\citep{libonati20213d, zolotovsky2021fish, shah2004modeling}, but there are a number of difficulties. In vivo 3D visualization and quantification of biological microstructural parameters are difficult to obtain as the tissue often changes during observation, and image segmentation techniques - needed to locate image features and boundaries (lines, curves, etc.) - carry errors and the process requires high (pixel level) accuracy and intensive manual intervention. A single image can take up to 30 minutes or more to process. With the development of image acquisition techniques such as microCT (microcomputed tomography) scans with lab-scale or synchrotron x-ray sources, and multiphoton microscopy it is possible to reach extremely high resolution (up to 300nm/voxel), but this results in large datasets which are not easy to handle. Medium resolution microCT scans  (6,25 $\mu$m/pixel) of a cylinder of meniscus with a diameter of 1mm and length 3mm, created an image dataset of over 0.3TB. Analysing these large datasets and quantifying distributions of architectural and material parameters is computationally expensive.

Recent advances from our group on image analysis methods~\citep{WaghornePerm,WaghorneTort} allow the statistical quantification of architectural features -porosity, pore dimension, pore connectivity and tortuosity -  of such natural and highly porous materials. However, the aim of this work is to lay the foundations for an innovative approach to revealing the content of this data whilst finding suitable lower order representations of it. The idea is to achieve this by moving through sensorial spaces, particularly sound, and exploring how the relationships between them can enhance our perception of the data. In particular we aim is to develop a transperceptual method to to detect architectural features which might be missed by the analysis of segmented images (Fig.~\ref{fig:scheme}).

Sonification is the presentation of data via the medium of sound, in contrast to the typical visual display of data (visualisation). Sonified data utilises aural processing to try and rapidly understand large data-sets which may be difficult or time-consuming to visualise and view on-screen~\cite{10.1117/12.2016019}, and is an appealing approach to enable data analysis by those who struggle to visually process data~\cite{doi:10.1073/pnas.1705325114,ballora_2014}. Sonification has previously been applied to data from fields as diverse as medicine,\cite{898630,Ahmad:10} astronomy~\cite{Zanella2022}, and chemistry~\cite{mahjour_bench_zhang_frazier_cernak_2022}, with many different approaches taken to map sound on to data and produce an audio output.

Challenges in the sonification field include identifying the best mappings of sound onto a data set to render subtle changes and differences audible, and building the tools to render numerical or visual data into sound quickly and efficiently. The key advancement in the method developed here is the adoption of a transperceptual approach~\cite{harris2021composing} to data exploration in which sound and image are used in combination and reciprocally as opposed to in isolation, which can be typical of standalone sonification projects. Research points to the value of sound and image in enhancing one another, and the aim of a transperceptual approach is to capture that added value~\cite{chion2019audio} and harness it for the purposes of data exploration.

To unify our audio and visual data we have also explored the use of audio-to-image and image-to-audio generative adversarial networks (GANs). GANs have achieved excellent performance in many domains, such as image synthesis~\cite{goodfellow2014generative,StyleGAN22019analyzing}, image to image translation~\cite{zhu2017unpaired,isola2017image}, audio to image generation~\cite{duarte2019wav2pix,wan2019towards}.
In this work, we utilize audio-to-image GANs to learn the relationship between audio and image data. We train the sound domain with the audio files of each image. More specifically, given a piece of audio, we utilize an audio-to-image GAN to generate the corresponding image. From the perspective of human perception, this method provides us with a way of analyzing audio data from vision data. Here, the potential reciprocity between image analysis, sonification and subsequent training of the GAN as part of a transperceptual approach offers the potential to expand the method to the point where only the GAN is needed to reverse engineer the initial image analysis. Although at an early stage, the results thus far are promising.

Here we start from samples of knee meniscus, and give an overview of the imaging data (micro-Ct scans) and explore image analysis techniques for the quantitative analysis of the architectural properties. We then apply a sonification processes to convert the data (images) into various sounds. These images and sounds were then used to train and test a GAN. Our results show that the 3D structure of complex tissues can be understood in the sound domain, and with our GAN we can successfully generate an artificial dataset of soft tissues with biomimetic architectures.

\section{Materials and Methods}
\subsection{Image acquisition, processing and statistical analysis}\label{sect-preprocessing}
\begin{figure}[h!]
   \centering
     \includegraphics[width=1\textwidth]{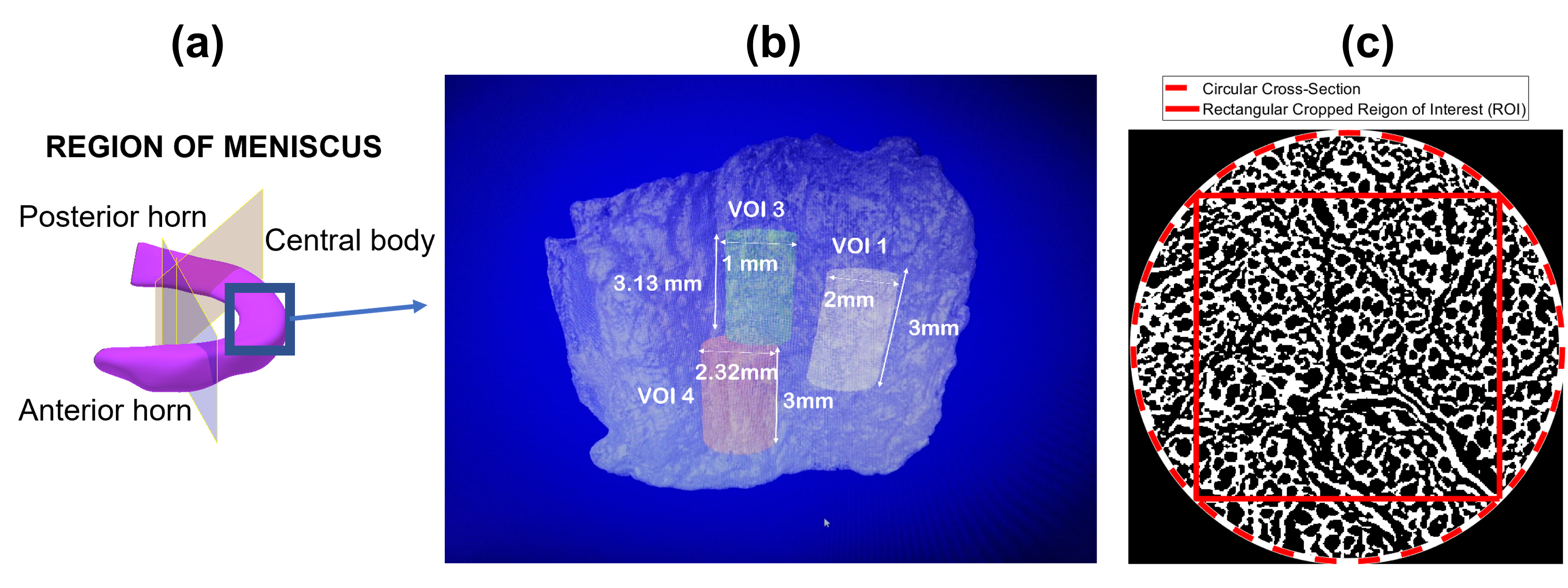}
       \caption{a. Schematic representation of lateral meniscus, b. The scanned portion of the central body of dimensions $9mm\times 10mm\times 12mm$ and the cylindrical dataset extracted. Here we analyse and mimic VOI4, c. An example of image from the 2D stack - in total 478 images reconstructed in 3D - of VOI4 of diameter 2.32mm}
      \label{fig:parametersStat}
\end{figure}

A sample of dimension $9mm\times 10mm\times 12mm$ (Fig.~\ref{fig:parametersStat}b) was extracted from the centre part of the medial meniscus - the body (Fig.~\ref{fig:parametersStat}a)-  with a surgical knife and immediately freeze dried in a Benchtop Freeze Dryer (FreeZone Triad Cascade, Labconco) following the procedure in~\citep{bonomo2020procedure}.
Micro–Computed Tomography ($\mu$CT) analyses were carried out with a $\mu$CT scanner (Skyscan 1272, Bruker Kontich, Belgium). The images were acquired with an image resolution of 6,25 microns. The scanning dataset obtained after the acquisition step consisted of images in 16-bit tiff format ($371 \times 371$ pixels). The acquired images were reconstructed by software N-Recon (version 1.6.10.2), 4 different Volumes of Interest (VOIs) were extracted from different regions of the scanned sample (Fig.~\ref{fig:parametersStat}b)~\citep{WaghornePerm}. We have analyzed in detail here VOI4 - a cylinder of diameter of 2.32mm and height of 3mm, in total this dataset contains 478 images. It was necessary to crop the circular cross-section of all images to rectangular due to the background pixels present on the scans that, which would impair the sonification. The dataset was cropped to an image size of ($256 \times 256$ pixels) as shown in (Fig.\ref{fig:parametersStat}c). From here two separate preprocessing techniques have been employed, with each method to be used for a different experiment. In the first instance, the images have compressed to fit within a $64 \times 64$ frame using the function 'Resize' from the  Pillow library~\cite{clark2015pillow} within python; this was due to a limitation of the GAN which will be discussed later. A workflow for this 'downsampled' data can be seen in (Fig.~\ref{fig:ds_preprocessing}) along with an example of of its generated  counterpart. As the original images have only been downsampled for this process, the number of images input remains at 478. For the second experiment, rather than downsampling the images to shrink them to the required size, the whole images are instead segmented into $64 \times 64$ squares. This method increases the data input to 7468 images and avoids the loss of any information due to downsampling, however, this comes at the expense of no longer interpreting the each sample holistically. As the samples are deconstructed in preprocessing, this means that the corresponding images generated needed to be reconstructed in post-processing in order to maintain it's spatial resolution, a workflow for this 'full resolution' dataset can be seen in (Fig.~\ref{fig:fr_preprocessing}).

\subsection{3-Dimensional analysis }
In order to quantify the characteristic features of the microstructures, a Matlab code was written for image processing of the 3D dataset and perform statistical analysis of the architecture\citep{WaghorneTort}. The main steps of the software are as follow: 
\begin{itemize}
\item The images are initially preprocessed by binarisation  using Ostu's method followed by a 'majority' transform  to removed insignificant features that can interfere with the segmentation \cite{Rabbani2014}  
\item A distance transform is performed on the binary 3D image along with with median filtering to improve the quality of segmentation. Watershed segmentation is then applied to this distance map, leading to the labelling and quantification of pores~\citep{Rabbani2014}. This approach is seen to consistently yield result similar to more classical methods at  a fraction of the computation time~\cite{Baychev2019}.
 \item Morphological characteristics can be acquired from the segmented pore space, most importantly pore size. The size is described by created an 'equivalent sphere' of equal volume to the pore, the diameter of the created sphere is what is used to describe it's size.
\item Using a methodology taken from~\cite{Rabbani2016}, the connectivity of the pore network can be identified. This is accomplished by dilation of the 3D binary image, in combination with the labelled pore matrix attained by segmentation, to interpret the connection of pores.
\item Lastly the pore network can be modelled as a graph, with the centroid of pores acting as nodes and edges representing the distance between connected pores. With inspiration taken from~\cite{Sobieski2016} regularly spaced start and finish points are assigned to the graph. The shortest path between all possible start and finish points is then calculated, providing a distribution of geometric tortuosities~\cite{Ghanbarian2013}.
 \end{itemize}

\subsection{Sonification}\label{sect-sonification}

The sonification involved the creation of a 255-voice synthesiser mapped to the luminance (brightness) of the pixels in the images. The process involved sorting each image in turn to assess the luminance of each pixel, then using the combined number of pixels at each luminance level to control the amplitude of the corresponding voice of the 255 voice synthesiser. The amplitude was scaled for each synthesiser voice to avoid overloading the audio buffer, however, this scaling was fixed for each voice as opposed to dynamic, with each voice being scaled in relation to the others. This model can run in realtime and did not require modification of the images, but also involved significant amounts of phase interference due to the proximity of the frequencies in the synthesiser. 
The model (at this stage), was unable to retain spatial information regarding the location of areas of luminance in the image. An initial approach to sonification involved downsampling the images to a smaller size (16 x 16) and conducting per-pixel, pitch-based sonification up to a maximum of 1024 voices. This had advantages with regard to retaining information concerning the spatial location of brightness within the image, but involved a reduction in the resolution of the images and the necessity to segment the original images to deal with the downsampling, which had implications for the subsequent training of the GAN. For the next stage of this project, it would be useful to explore the possibility of dynamic scale as an alternate approach, in order to more accurately represent the preponderance of luminance values within the image and retain some spatial information.

In the downsampled ($64 \times 64$ pixels) images there was limited relationship noted between the original images and those generated by the GAN. Consequently, the 478 images at full resolution were segmented into 478 sets of 16 images, with sonification then performed on these segmented images - a total of 7648 images, yielding the same number of audio files which were subsequently passed to the GAN.

The produced audio datafiles are available as supplementary data, and clearly display phase interference/cancellation. In future we hope to use this interference effect as an element of the sonification, and develop the system to include spatial information. 

\subsection{Audio-image GANs} 

\begin{figure}[h]
\centering
\includegraphics[scale=0.5]{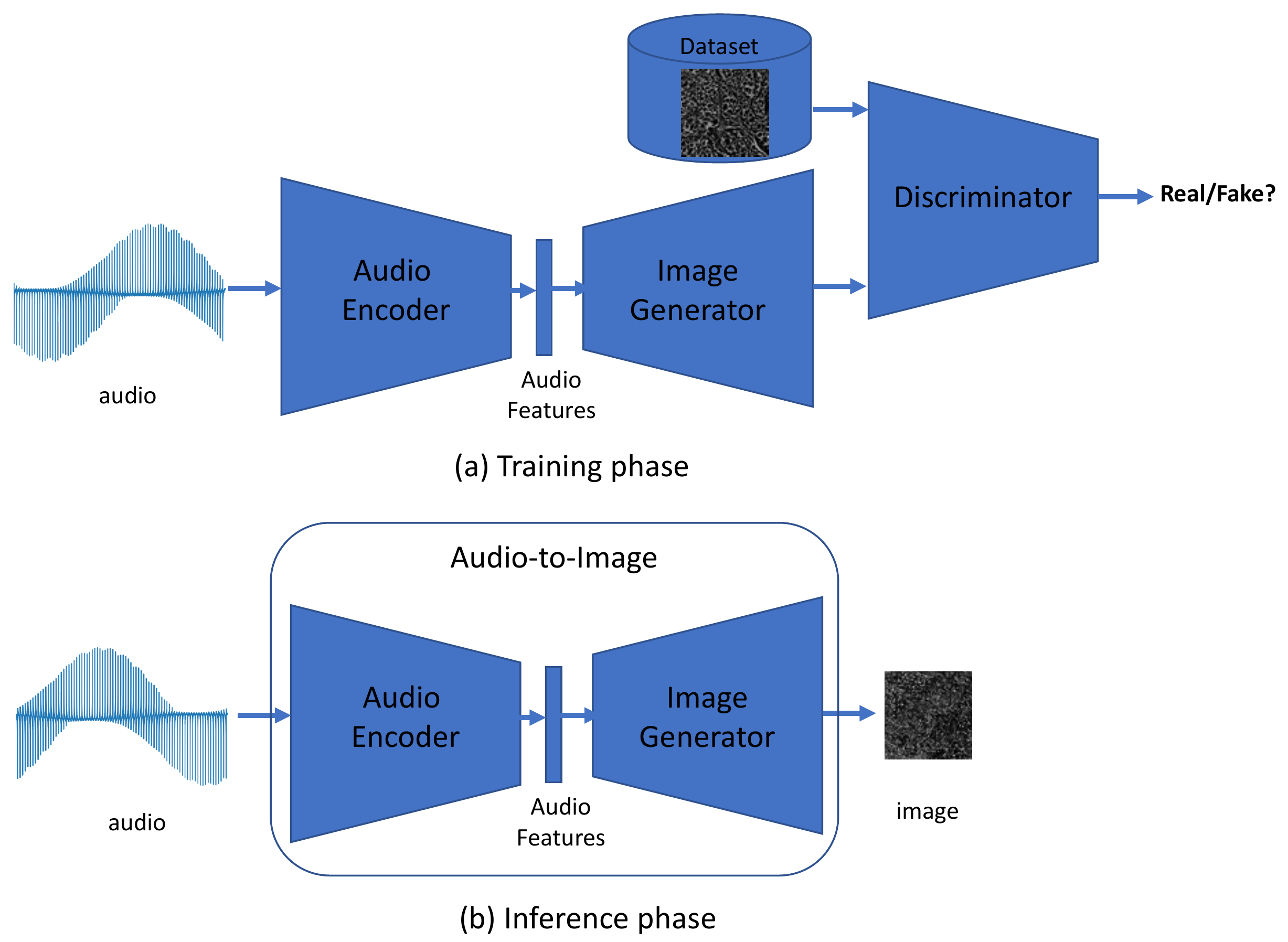}
\caption{The audio-image GAN architecture. In the training phase, audio-image pairs are used for training. In the inference phase, given a piece of audio, generated images can be obtained through the audio-to-image GAN.}
\label{fig:gan_architecture}
\end{figure}

We utilize the audio-image generative adversarial network~(GAN) proposed by~\cite{duarte2019wav2pix} to model the relationship between audio and image data.
(Fig.~\ref{fig:gan_architecture}) shows the training and inference phase for the audio-image GAN.
In the training phase, it mainly consists of three neural network modules: an audio encoder, an image generator, and a discriminator.
The audio encoder takes a piece of audio as an input, and generates corresponding audio features.
The audio features are fed into the image generator and a fake image can be generated.
The discriminator is responsible for differentiating the fake image from the generator and the real image from the training set.
After finishing the training process, only the audio encoder and the image generator are utilized in the inference phase.
Specifically, given any a piece of audio, the audio encoder transforms it as audio features. Then a generated image can be obtained through the image generator.

\begin{figure}[h]
\centering
\includegraphics[scale=0.5]{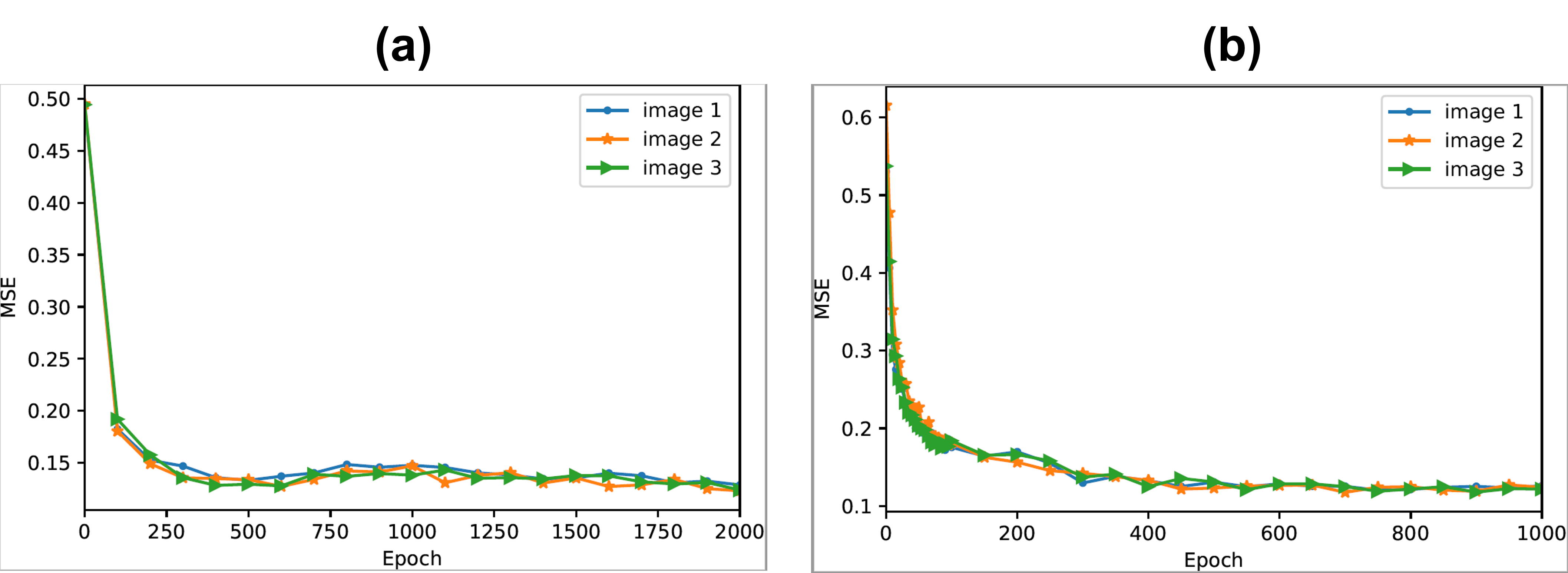}
\caption{Reduction of MSE during the GAN training process. (a) Downsampled dataset, (b) full-resolution dataset}
\label{fig:gan_architectureMSE}
\end{figure}

The initial dataset contained 478 pairs of image and audio data.
We randomly chose 400 pairs of data as training data and the remaining data was used for testing. For the larger segmented image dataset, we randomly choose 7000 pairs of data as training data and 648 data pairs for testing.
We have based our approach on an implementation of work by~\cite{duarte2019wav2pix} for the audio-image GAN.
Specifically, we set the number of channels as 1 because in our work images are grayscale.
The number of epochs (learning iterations) is set as 2,000 for a GAN model trained on downsampled images and 1000 for that on the full resolution images.
The learning rate of the generator and the discriminator
is fixed as 0.00001 and 0.00004. In (Fig.\ref{fig:gan_architectureMSE}) we show that the MSE reduces more rapidly in the case of the full-resolution dataset -(Fig.\ref{fig:gan_architectureMSE}b) - than in the case of the downsampled dataset shown in (Fig.\ref{fig:gan_architectureMSE}a). Therefore the learning iterations (epoch) for the full resolution image training set has been set to 1000.

\section{Results and Discussion}
 The sonification process outputs (audio files) extracted using the methods described in section \ref{sect-sonification} from both these experiments, along with their corresponding images, are provided in the Supplementary Material section. The audio-image GAN resulted in the images shown in  Figures (\ref{fig:ds_preprocessing}c) and (\ref{fig:fr_preprocessing}d). We use mean squared error~(MSE) to measure the difference between original images and generated images to evaluate the quality of reconstructed images.
Given original images $X_{\it ori} = \{ x_{\it ori}^1, x_{\it ori}^2,..., x_{\it ori}^n\}$, the corresponding reconstructed images $X_{\it res} = \{ x_{\it res}^1, x_{\it res}^2,..., x_{\it res}^n\}$, a MSE is computed by: 
\begin{equation}
	{\it MSE} = \frac{1}{n} \sum_{i=1}^{n} (x_{\it res}^i - x_{\it res}^i)^2  
	\label{eq:mse}
\end{equation} 

\begin{figure}[h]
	\centering
	\subfigure[Downsampled dataset.]{	\includegraphics[width=0.45\columnwidth]{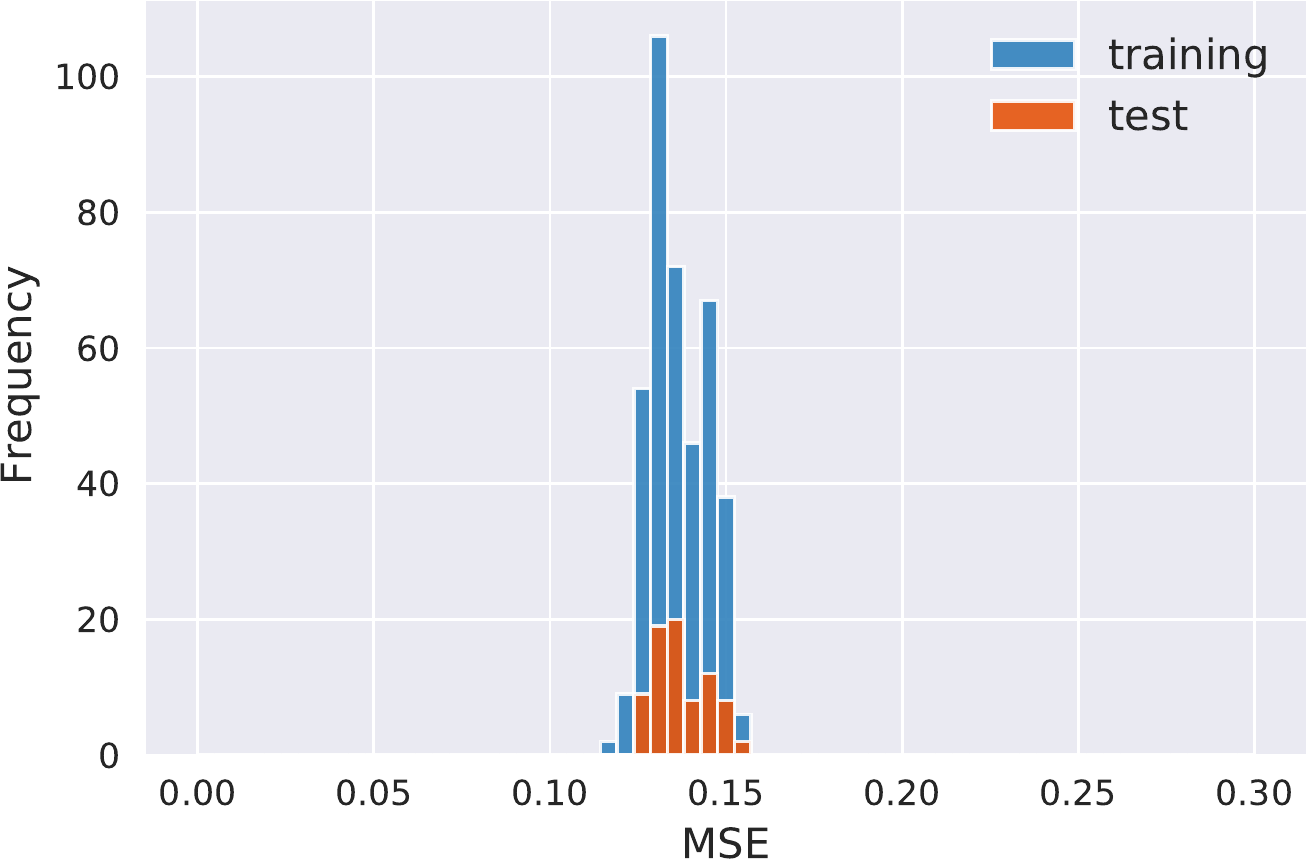}
		\label{fig:errors_downsamp}
	}
	\subfigure[Full resolution dataset.]{	\includegraphics[width=0.45\columnwidth]{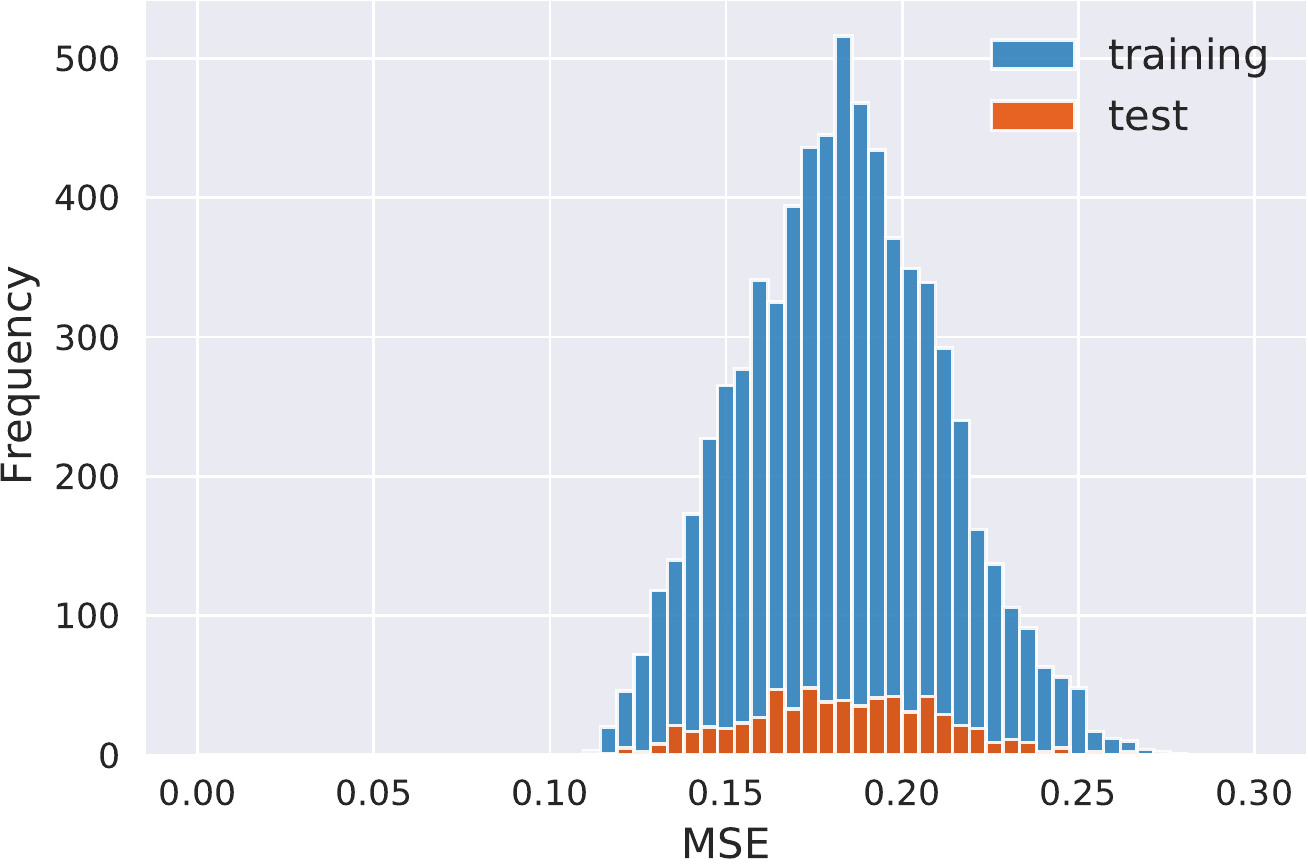}
		\label{fig:errors_seg}
	}	
	
	\caption{MSE of training and test data.}
	\label{fig:errors}
\end{figure}

(Fig.~\ref{fig:errors}) shows the frequency of MSE on training and test data. 
We can observe that the MSE distribution of training and test data is similar.
Specifically, the mean MSE of training data is 0.1365 while the test data has a mean MSE of 0.1372 for the downsampled dataset.
The full resolution dataset has a mean MSE of training data of 0.1831 and the mean MSE of test data of 0.1829.
(Fig.~\ref{fig:ds_preprocessing}) shows some examples for original images and reconstructed images.
Overall, given audio samples as input, the GAN  can roughly predict the trends of the original images.

We ran image analysis on both the 2D stack and the reconstructed 3D objects in order to quantitatively compare the statistical distribution of architectural parameters (porosity, pore-size, pore connectivity and tortuosity) of the generated versus original datasets.

\subsubsection*{Downsampled Dataset: 478 pairs of images and audio}
\begin{figure}[h]
	\centering
	\includegraphics[width=1\textwidth]{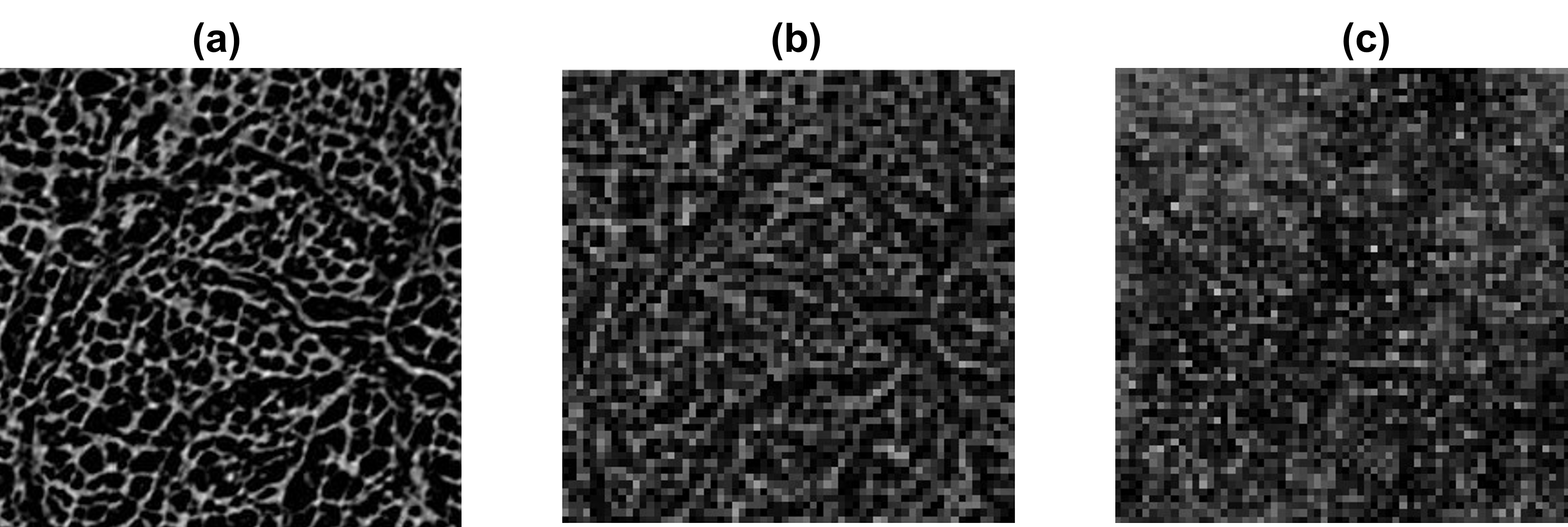}
	\caption{Workflow of downsampling analysis: (a) Original dataset (256 $\times$ 256) , (b) Downsampled image (64 $\times$ 64), (c) GAN generated image (64 $\times$ 64)}
	\label{fig:ds_preprocessing}
\end{figure}
\begin{figure}[h]
	\centering
	\includegraphics[width=1\textwidth]{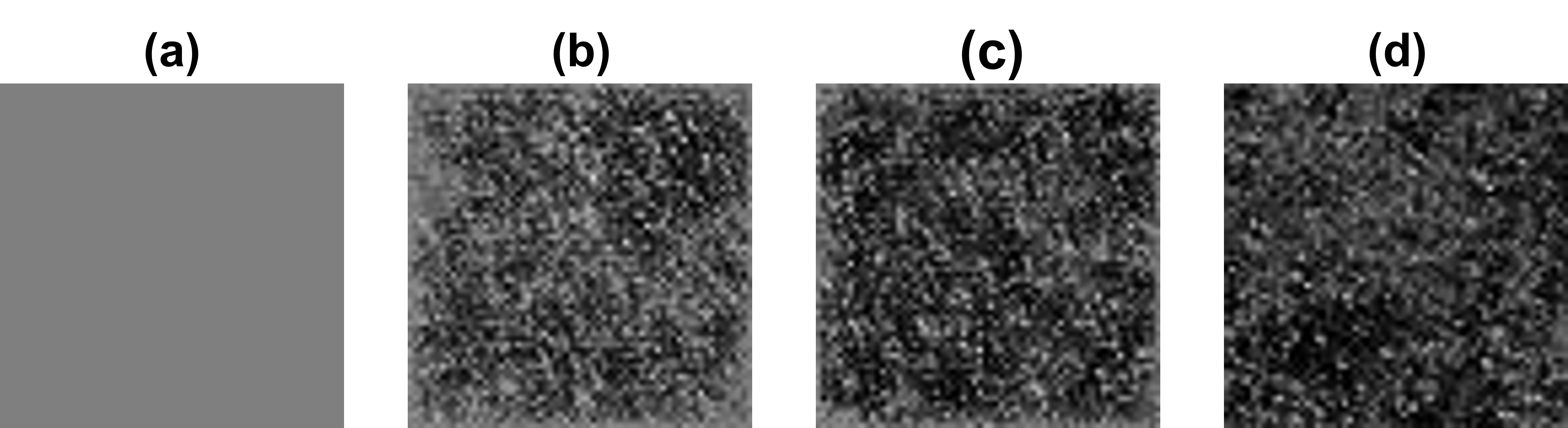}
	\caption{Training progression of GAN with downsampled images. (a) learning iteration=0, (b) learning iteration=50, (c) learning iteration=100,(d) learning iteration=2000}
	\label{fig:GAN_learn_ds}
\end{figure}
\begin{figure}[h]
	\centering
	\includegraphics[width=\textwidth]{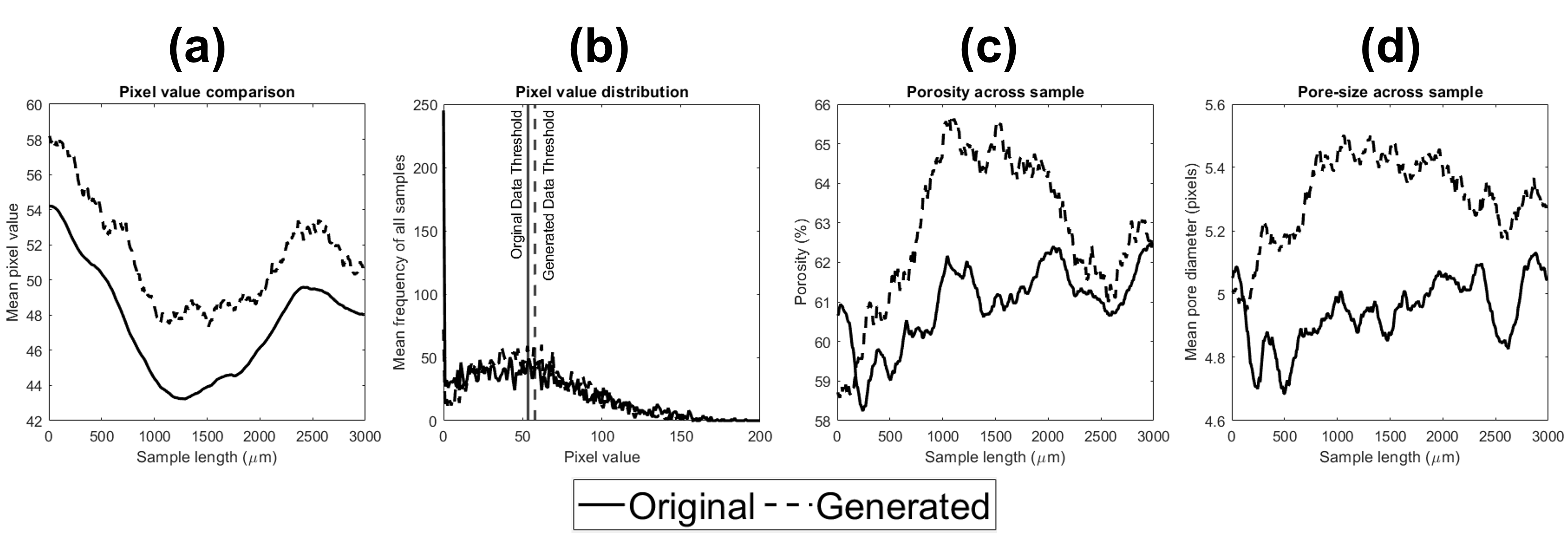}
	\caption{2D Analysis of downsampled dataset: (a) Mean luminance value per image, (b) Luminance distribution, (c) Porosity across sample, (d) Pore size across sample}
	\label{fig:anlys_ds}
\end{figure}

As discussed in section \ref{sect-preprocessing} first attempt at producing mimetic native tissues was met by means of downsampling the original images down to fit within a 64 $\times$ 64 frame, as seen in (Fig.\ref{fig:ds_preprocessing}). Before analysing the results produced from the GAN, we show the learning process of the GAN. Before any information has been passed through the GAN the image produced appears as solid gray (Fig.\ref{fig:GAN_learn_ds}a). At the start of the learning process, the distribution of luminance values seem to closer to the original image, the pixels are however, seemingly spread randomly distributed across the image frame as seen in (Fig.\ref{fig:GAN_learn_ds}b). By the end of the process, the GAN appears to have learnt some of the key characteristic of the images, most noticeably clustering. The final images in the training stage (Fig.\ref{fig:GAN_learn_ds}c,d) demonstrates an image that appear more frequently in pockets of darker pixels. This is the GAN's method of emulating pores, not because there has been any spatial information provided by the audio, but because having clusters of dark regions increase the chance of producing a lower MSE by matching with pores.  
These datasets have been analysed first in the 2D, highlighting the discrepancy in parameters between corresponding original and generated image pairs. The parameters evaluated are first related to the information provided to the GAN via the audio, in this instance quantities of pixel luminance in (Fig.~\ref{fig:anlys_ds}a,b). (Fig.~\ref{fig:anlys_ds}a) is created by quite simply calculating the mean average of all the luminance values within an image. It is clear that the generated images stably follow the same trend as the native dataset, on average there being an 8\% difference between the images. It is worth noting that, for legibility, a moving average window of 25 has been applied to (Fig.~\ref{fig:anlys_ds}a,c\&d); this is also true for (Fig.~\ref{fig:fr_anlys}). (Fig~\ref{fig:anlys_ds}b) shows the distribution of the average luminance values and demonstrate good agreement between them, in both range and values, with the exception 0 luminance. The model appears to have difficulty recognising totally black pixels, this is likely to be due to the relative scaling of amplitudes during the sonification process. There is a large discrepancy between the quantity of black pixels and all others gray scale pixels. This difficulty interpreting black pixels is also a contribution to the offset in average image luminance seen in (Fig.~\ref{fig:anlys_ds}a).
The second set of results, (Fig.~\ref{fig:anlys_ds}c,d) require some post processing before it is possible to analyse. The filtering and segmentation methodology laid out for 2D images in~\cite{Rabbani2014} was followed to provide a quantified 2D pore space, it was found that a [2 2] median filter produced the best results. The porosity seen in (Fig.~\ref{fig:anlys_ds}c) also has very good agree-ability, with on average there being a 4\% difference between the values. Interestingly it is seen that the GAN tends to create images that overpredict porosity, despite having higher average luminance values, this is likely due to the binarisation method used. While the similar distributions produce a similar average threshold value, as seen in (Fig~\ref{fig:anlys_ds}b), the threshold of the generate images is slightly higher, this will lead to more frequent black pixels and a higher porosity, this also explains why the porosity tends to be better in regions where the mean luminance value is closer to the threshold. Lastly, in (Fig.~\ref{fig:anlys_ds}c) we show the average pore size. While is demonstrates an averages an 8\% difference in pore diameters, this parameter requires spatial information to successfully match. The resemblance in trends and close values we see has been entirely learned by the GAN network by trying to minimise then MSE. The results show an overestimation, likely because having a slightly larger cluster of dark pixels allowing for the crossover of more than one pore. The large clusters of dark pixels would likely decrease the MSE as there are more black pixels present on the original data.  

\subsubsection*{Full resolution dataset: 7648 pairs of images and audio}
\begin{figure}[h]
	\centering
	\includegraphics[width=1\textwidth]{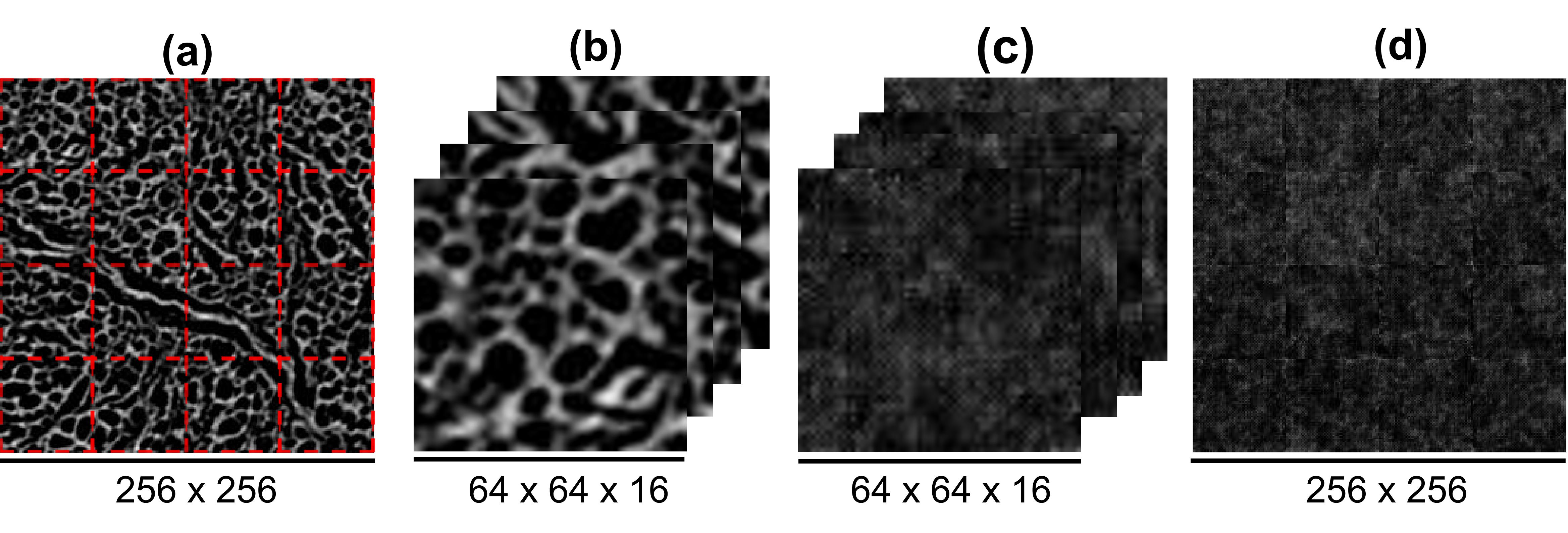}
	\caption{Workflow of full resolution analysis: (a) original image showing segmentation lines (red), (b) deconstructed original image (c), deconstructed generated image, (d) reconstructed generated image}
	\label{fig:fr_preprocessing}
\end{figure}
\begin{figure}[h]
	\centering
	\includegraphics[width=1\textwidth]{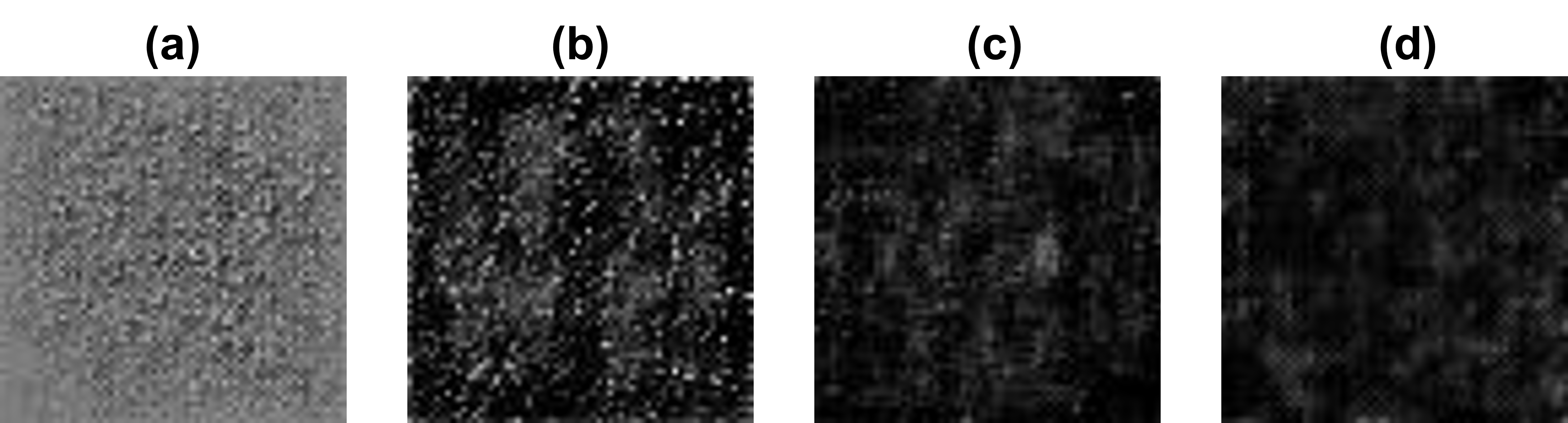}
	\caption{Training for full resolution dataset. (a) learning iteration=5, (b) learning iteration=100, (c) learning iteration=750, (d) learning iteration=1000  }
	\label{fig:fr_train}
\end{figure}

The second experiment revolves around maintaining as much of the native morphology as possible by segmenting the initial full resolution images (256 $\times$ 256), rather than downsampling them. This experiment comes with additional step of reconstruction for analysis as seen in (Fig.~\ref{fig:fr_preprocessing}c,d). The additional step consists of segmenting the image into 16 squares of size (64 $\times$ 64)  assigning unique names for each segmented image based on the location in the image stack and maintaining this through all processing stages. Barring the reconstruction, all other aspects of the analysis remained the same as the downsampled dataset. 
The first step, again, was be to interpret the GAN's learning process to ascertain an understanding of what additional features have been detected compared to the downsampled dataset. The learning process begins relatively similar to that of the downsampled, with (Fig.~\ref{fig:fr_train}a,b) demonstrating the formation of clusters,however, towards the end of the training the images seem to take on a different form. (Fig.~\ref{fig:fr_train}c,d) present blurring within some sections on of the generate image and a seemingly cross-hatch pattern in others. One of the possible reason is that, due to retaining the original resolution of the image, there are more frequent black pixels (luminance=0) and more intensely bright pixels. The GAN's way of minimising the MSE is to not include clusters or distinct boundaries, but instead to have smoother areas with less intense luminance values. This avoids to incur a high MSE if pixels were to fall within the wrong side of the pore boundary.

\begin{figure}[h]
	\centering
	\includegraphics[width=\textwidth]{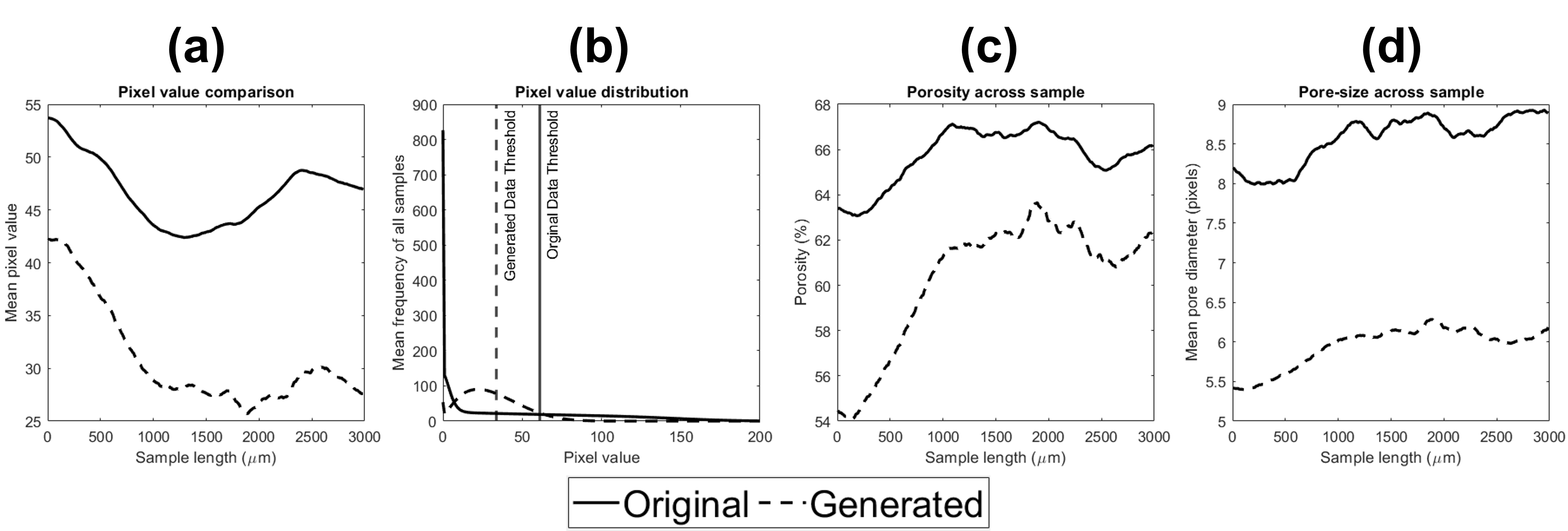}
	\caption{2D Analysis of full resolution dataset: (a) Mean luminance value per image, (b) Luminance distribution, (c) Porosity across sample, (d) Pore size across sample}
	\label{fig:fr_anlys}
\end{figure}

The first two results of(Fig.~\ref{fig:fr_anlys}) support this theory, as (Fig.~\ref{fig:fr_anlys}b) shows a clear shift of the GAN's luminance distribution to the left, with hardly any pixels with luminance $>$ 75, this is to accomodate for the now huge discrepancy of zero luminance values to all others in the original data. This shift is reflected within  (Fig.~\ref{fig:fr_anlys}a), now with the GAN consistently producing pixels skewed to be darker, the average luminance value drops considerably, even lower than the naturally morphology albeit with it's spike in 0 pixels. The difference mean luminance value in this experiment averages almost 4$\times$ that of the previous. The Porosity in this experiment is not only considerably different in values but as (Fig.~\ref{fig:fr_anlys}c) shows, the GAN is now underestimating the porosity, rather than overestimating. This again is due to the binarisation process, as the distributions of the luminance values are very dissimilar. The original data has clearly defined background and foreground pixels with it's much more contrasted range, while the generated data is not so dichotomic. This causes the threshold for the generated to be much lower than that of the original. This discrepancy in threshold along with original data's heavy right skew explains the large discrepancy in porosity. The pore size comparison does not perform well in this experiment, the GAN does not emphasise clustering. Therefore the pore size is considerably lower than the native morphology, on average being 30\% lower. 

\begin{figure}[h]
	\centering
	\includegraphics[width=\textwidth]{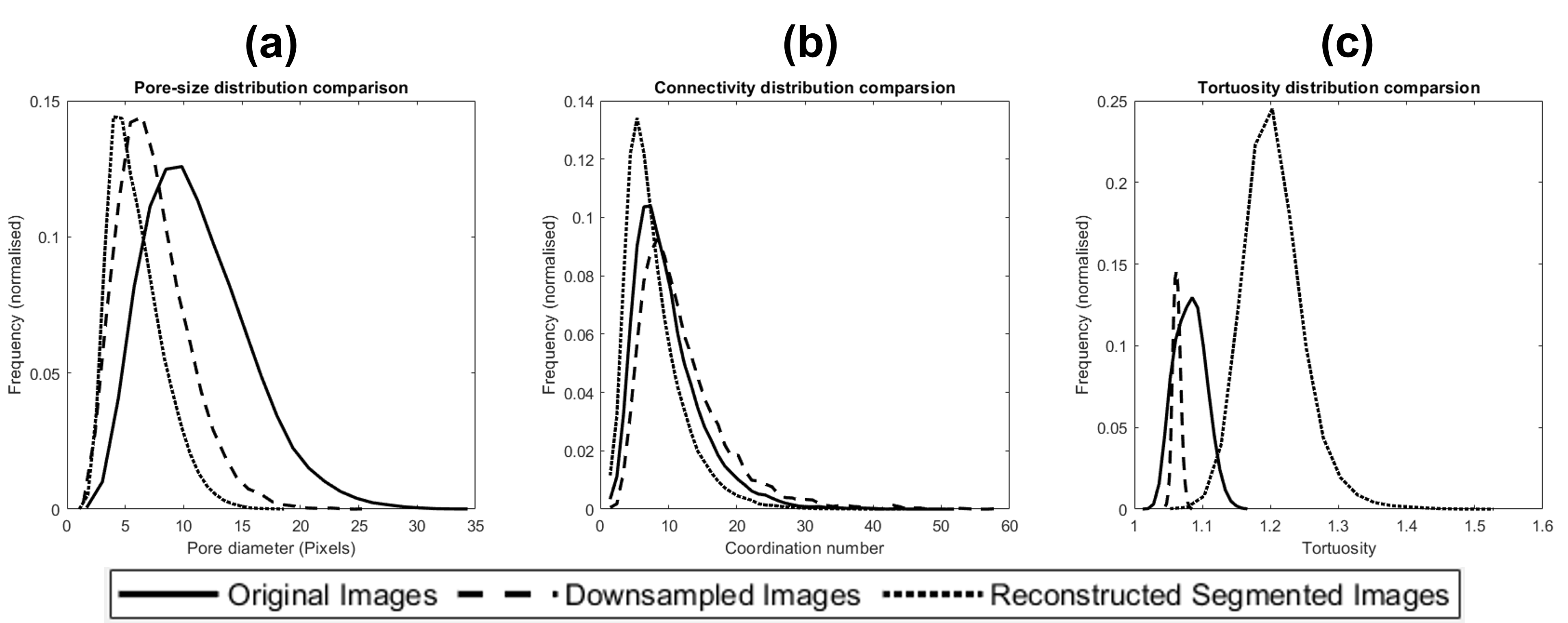}
	\caption{3D Analysis of all datasets: (a) 3D Pore size distribution, (b) Pore connectivity distribution, (c) Tortuosity distribution}
	\label{fig:3d_anlys}
\end{figure}

 The results of the 3D  analysis describe the effect of how the GAN model learning process can affect the generation of images. It is first important to note that these graphs are all normalised on the Y-axis this is because we are interpreting images of varying size and this will have a large effect on the frequency of occurrence. The results have been normalised to be compared. In (Fig.~\ref{fig:3d_anlys}a), We can see the variation in distributions of the pore size. The largest and widest distribution belongs to the original, with spatial relation of these pores  larger in 3D space. This is followed closely by the downsampled dataset which is  only 1/4 the size of the original one. If the resolution of the image were to be scaled up, it would like have clusters considerably larger than the natural morphology. Lastly as very little clustering occurs in the full resolution dataset, further intensified by a lack of realization between planes, this results in a very thin pore size distribution. In (Fig.~\ref{fig:3d_anlys}b) the distribution of pore connectivity appears relatively similar among the original, downsampled and full scale images. It is seen that there is a relationship between the pore size and connectivity due to it's available surface area. This is likely the reason for the full resolution dataset presenting the thinnest distribution. The downsampled and original connectivity are relatively close, with downsampled seemingly having a wider spread, this is perhaps due to the lack of restriction in pore size which in the natural morphology are given by a net contrast with the solid matter around pores. 
 Tortuosity - (Fig.~\ref{fig:3d_anlys}c) - is visibly where the analysis differs the most, the downsampled dataset is likely highly limited by the reduced size, there is likely far fewer start and finish points and the channel is restricted to a very small area, leading to smaller tortuosities. The full resolution dataset however, presents smaller pores and less connections, therefore it leads to a more uniform tortuosity and to the wider distribution in (Fig.~\ref{fig:3d_anlys}c).

\section{Conclusions and Future Perspectives}

From the results presented we have demonstrated that it is in fact possible for a GAN to produce an artificial image dataset following the parameter described to it through audio information. Not only this, but other characteristics can also be self taught by during by the GAN through learning process. While the audio information and results presented so far may seem primitive, this lays the foundation for a ramping in complexity. With audio information that can describe more features, not only will results improve but also the potential for the GAN to uncover a hidden characteristic not clearly visible with visual perception alone. These kinds of developments could be used in a wide range of application such as assisting with diagnoses, exploration into new materials and generation of artificial structures to name a few. 

This investigation has also highlighted some potential weaknesses with the methodology. The interaction between the audio information and GAN is complicated and difficult to evaluate directly. A standardised methodology for evaluating weighting factors of the GAN's neural network would be highly beneficial for improving results and uncovering relationships faster. To further this, careful thought and consideration must go into selecting what data would be most suitable for use. In these experiments two different sets of data were used, one containing downsampled images and the other full resolution data with a unit of magnitude greater volume of data, with the latter set performing considerably worse. This is a paradigm example how situations that may typically look good in theory, do not behave so in practicality. If, however, these precautions are made then this methodology has the potential to be very powerful. 

The method proposed in this paper has the potential to be adopted for analysing and predicting a wide variety of biomimetic architectures. The original dataset analysed consisted of a stack of 478 circular 2D images ($371 \times 371$ pixels) which have been reconstructed as a 3D object whose architectural properties have been statistically characterized, sonified and then reconstructed. 
2D and 3D analysis of the reconsructed 2D stack of images shows that the audio visual GAN performed better in generating artificial dataset when trained on the downsampled images. The mean of the luminance values, porosity and pore size are within 4-8percent of the original dataset.

The proposed method has the following limitations:
\begin{itemize}
    \item The current GAN architecture can only deal with the image sizes up to $64 \times 64$ pixels. The sonification process is able to deal with any size of image.
    \item The sonification method currently can underestimates the sound of black pixels (pore space).The downsampled dataset contains less black pixels than the full resolution images. So, however the images to train the GAN are cleared when using the full scale images, the actual sonification methodology is more suitable for the downsampled dataset. 
\end{itemize}

Our future development will focus on how to use the GAN to more automatically generate audio datasets from the sample scans, and investigate how audio data might be perceived differently for different samples by humans as well as computers.

\section*{Acknowledgements}
The authors are grateful to the financial support provided by the FNR-funded European Crucible Prize.

%\newpage
\bibliographystyle{elsarticle-num}
\bibliography{references}

\end{document}